# Incremental Recompilation of Knowledge

**Goran Gogic**                                                              GORAN@CS.UCSD.EDU
*University of California at San Diego*
*Computer Science Department*
*La Jolla, CA 92039-0114*

**Christos H. Papadimitriou**                                    CHRISTOS@CS.BERKELEY.EDU
*University of California at Berkeley*
*Soda Hall 689, EECS Department*
*Berkeley, CA 94720, U.S.A.*

**Martha Sideri**                                                                  MSS@AUEB.GR
*Athens University of Economics and Business*
*Patission Street, Department of Informatics*
*Athens 104 34, Greece*

## Abstract

Approximating a general formula from above and below by Horn formulas (its Horn envelope and Horn core, respectively) was proposed by Selman and Kautz (1991, 1996) as a form of "knowledge compilation," supporting rapid approximate reasoning; on the negative side, this scheme is static in that it supports no updates, and has certain complexity drawbacks pointed out by Kavvadias, Papadimitriou and Sideri (1993). On the other hand, the many frameworks and schemes proposed in the literature for theory update and revision are plagued by serious complexity-theoretic impediments, even in the Horn case, as was pointed out by Eiter and Gottlob (1992), and is further demonstrated in the present paper. More fundamentally, these schemes are not inductive, in that they may lose in a single update any positive properties of the represented sets of formulas (small size, Horn structure, etc.). In this paper we propose a new scheme, *incremental recompilation*, which combines Horn approximation and model-based updates; this scheme is inductive and very efficient, free of the problems facing its constituents. A set of formulas is represented by an upper and lower Horn approximation. To update, we replace the upper Horn formula by the Horn envelope of its minimum-change update, and similarly the lower one by the Horn core of its update; the key fact which enables this scheme is that *Horn envelopes and cores are easy to compute when the underlying formula is the result of a minimum-change update of a Horn formula by a clause.* We conjecture that efficient algorithms are possible for more complex updates.

## 1. Introduction

Starting with the ideas of Levesque (1986) in recent years there has been increasing interest in computational models for *rapid approximate reasoning*, based on a "vivid" (that is to say, conducive to efficient deductions) representation of knowledge. One important proposal in this regard has been the *knowledge compilation* idea of Selman and Kautz (1991, 1996), whereby a propositional formula is represented by its optimal upper (relaxed) and lower (strict) approximations by Horn formulas —the corresponding Horn formulas are called in the present paper the *Horn envelope* and the *Horn core* of the original formula. The key idea of course is that, since these approximate theories are Horn, one can use them for rapid (linear-time) approximate reasoning.

Despite the computational advantages and attractiveness of this idea, some obstacles to its implementation have been pointed out. First, as was noted by Selman and Kautz (1991, 1996), the Horn approximations are hard to compute in general, and may in fact be exponentially large when compared to the formula being approximated. Second, although the Horn envelope of a formula is unique up to equivalence, the Horn core is not; that is, there may be exponentially many inequivalent most relaxed Horn formulas implying the given one. As was proved by Kavvadias, Papadimitriou and Sideri (1993), selecting the one with the largest set of models, or one that is approximately optimal in this respect (within *any* bounded ratio), is NP-hard. Another disadvantage is that the Horn envelope may have to be exponentially larger, as a Boolean formula, than the given formula. What is more alarming is that, even if the Horn envelope is small, it may take exponential time to produce. Even if we are given the set of models of the original formula, there is no known *output-polynomial* algorithm for producing all clauses of the Horn envelope. (An algorithm is output-polynomial if it runs in time that is polynomial *in both the size of its input and its output;* this novel and little-studied concept of tractability —and, unfortunately, related concepts of *intractability*— have proved very relevant to various aspects of AI.) In fact, it was shown by Kavvadias, Papadimitriou and Sideri (1993) that generating the Horn envelope from the models of a formula is what we call in the present paper *TRANSVERSAL-hard*, suggesting that it is problematic whether it has an output-polynomial algorithm. These negative complexity results for knowledge compilation (admittedly, quite mild when compared with the serious obstacles to other approaches to knowledge representation and common-sense reasoning, e.g., Eiter & Gottlob 1992; 1993) are summarized without proof in Theorem 1.

Our knowledge about the world changes dynamically —and the world itself changes as well. The knowledge compilation proposal contains no provisions for incorporating such belief revisions or updates. There are, of course, in the literature many formalisms for updating and revising knowledge bases and databases with incomplete information (Dalal 1988; Satoh 1988; Borgida 1985; Weber 1985; Ginsberg 1986; Eiter & Gottlob 1992; Fagin, Ullman & Vardi 1983; Winslett 1988; Winslett 1990; Forbus 1989). As was established by Eiter and Gottlob (1992), all these systems are plagued with tremendous complexity obstacles —even making the next inference, which is known as the *counterfactual problem*, is complete at some high level of the polynomial hierarchy for all of them. We point out in this paper (Theorem 2) some serious problems associated with computing the updated/revised formula in the two formula-based frameworks (Ginsberg 1986; Fagin, Ullman & Vardi 1983; Winslett 1988) *even if the formula being updated is Horn*. The only ray of hope from Eiter and Gottlob (1992) —namely that when the formula is Horn, the update/revision is small, and the approach is any one of the model-based ones, then counterfactuals are easy— is tarnished by the observation that, in all these cases, the updated/revised formula is not Horn (this is part (iii) of Theorem 2); hence, such an update/revision scheme would fail to be *inductive*, that is, does not retain its positive computational properties in the face of change.

To summarize, knowledge compilation of arbitrary formulas is not easy to do. And all known approaches to the update/revision problem encounter serious complexity obstacles, or result in loss of the Horn property. What hope is there then for a system that supports both rapid approximate reasoning *and* updates/revisions?

*Quite surprisingly, combining these two ideas, both shackled as they are by complexity-theoretic obstacles, seems to remove the obstacles from both, thus solving the combined problem*, at least in some interesting cases which were heretofore believed to be intractable. In particular we propose the following scheme: Suppose that formula $\Gamma$ is represented by its Horn envelope $\overline{\Gamma}$ and its Horn core $\underline{\Gamma}$ (to start the process, we incur a one-time computational cost for computing these bounds; alternatively, we may insist that we start with a Horn formula, in which case initially $\Gamma = \overline{\Gamma} = \underline{\Gamma}$). Suppose now that we update/revise our formula by $\phi$, a "simple enough" formula (how "simple" it has to be for our scheme to be efficient is an important issue which we have only partially explored; we know how to handle a single Horn clause, as well as several other special cases). *We represent the updated formula by the two formulas $\overline{\overline{\Gamma} + \phi}$ and $\underline{\Gamma} + \phi$*, where '+' stands for an appropriate model-based update/revision formalism. That is, our updated/revised upper and lower bounds are the Horn envelope of the updated upper bound and the Horn core of the updated lower bound. These are our new $\overline{\Gamma}$ and $\underline{\Gamma}$. In other words, we update/revise the two approximations, and approximate the two results, each in the safe direction. And so on, starting from the new approximations. The key technical point which makes this scheme work is that, although updating/revising Horn formulas, even by Horn clauses, does not preserve the Horn property, and finding Horn envelopes and cores is hard in general, *it is easy when the formula to be approximated is the result of the update/revision of a Horn formula by a Horn clause*. To our knowledge, our proposal, with all its restrictions, is the first computationally feasible approach to knowledge approximation *and* updates.

As the following example suggests, our proposal exhibits a desirable and intuitively expected "minimum-change" behavior, best demonstrated in the case in which a Horn formula $\Gamma$ is updated by a Horn clause, say $\phi = (x\&y \rightarrow z)$. Suppose that $\Gamma$ can be written as $x\&y\&\neg z\&\Gamma'$, where $\Gamma'$ does not involve $x$, $y$, or $z$ —if this is not possible, that is to say, if $\Gamma$ does not contradict $\phi$, then $\Gamma + \phi = \Gamma\&\phi$. Then the upper and lower approximations are these: $\overline{\Gamma + \phi}$ is $(x\&y \leftrightarrow z)\&\Gamma'$, while $\underline{\Gamma + \phi}$ is $x\&(y \leftrightarrow z)\&\Gamma'$ (or $y\&(x \leftrightarrow z)\&\Gamma'$, recall that cores are not unique). Notice the attractive "circumscriptive" nature of the updates (resulting from the minimum-change update and revision formalisms that we are using).

We conclude this introduction with a few disclaimers. As should be expected, and is pointed out in this paper, the computational feasibility of our approach comes with a "semantic price:" The upper and lower bounds $\overline{\Gamma}$ and $\underline{\Gamma}$ do not correspond in any natural way to some formula $\Gamma$; in fact, depending on the update formalism adopted, $\underline{\Gamma}$ may even fail to imply, as might be expected, $\overline{\Gamma}$ (see Theorem 5 and the example that follows). The pair $(\underline{\Gamma}, \overline{\Gamma})$ should be most accurately understood not as an efficient approximation of knowledge revision and updates, or an efficient dynamization of knowledge compilation, but instead as a new, combined, efficient approach to *both* the problems of vivid and dynamic knowledge. Its effectiveness as a knowledge representation formalism (that is, its semantic proximity to the situations being modeled, updated, and approximated, especially after a large number of updates) can only be tested experimentally, by applying it to typical or classical knowledge representation problems. Its apparent advantages are that (1) it comes with efficiency guarantees, and (2) it addresses —but of course does not provably solve— both dynamic and approximation aspects of the knowledge representation problem. Also, despite the fact that our approach produces the next representation in linear time, there is no guarantee that the bounds will not become exponentially larger than the formulas handled or than necessary (for example, after repeated doublings of the size of the representation). Finally, although we do argue that our approach is surrounded by negative complexity results in almost all directions, we are of course not claiming that it is the only computationally feasible approach that is possible.

## 2. Negative Results

Let $\Gamma$ be a *propositional* formula. Define (Selman and Kautz, 1991, 1996) its *Horn envelope* $\overline{\Gamma}$ to be a Horn formula such that (a) $\Gamma \models \overline{\Gamma}$, and (b) there is no other Horn formula $\Gamma' \not\equiv \overline{\Gamma}$ such that $\Gamma \models \Gamma' \models \overline{\Gamma}$; that is, $\overline{\Gamma}$ is the strongest Horn formula implied by $\Gamma$; it is called the *least Horn upper bound* by Selman and Kautz (1991). Symmetrically, the *Horn core* to be the weakest Horn formula implying $\Gamma$ (it is called the greatest Horn lower bound by Selman and Kautz, 1991). Naturally, one could not hope that the Horn envelope and core can be efficiently computed for all Boolean formulas. The reason is simple: $\Gamma$ is unsatisfiable iff both $\overline{\Gamma}$ and $\underline{\Gamma}$ are unsatisfiable —and it is well known that Horn formulae can be checked for satisfiability in linear time. But what if $\Gamma$ is given in some more convenient form, say in terms of its set of models $\mu(\Gamma)$ (that is, in "full disjunctive form")? A first problem is that $\overline{\Gamma}$ may have exponentially many clauses with respect to the size of $\mu(\Gamma)$ —there is little that can be done in this, we need them all to best approximate our formula. But can we hope to output these clauses, however many they may be, in time polynomial both in the size of input —$\mu(\Gamma)$— and of the output —$\overline{\Gamma}$? There are systematic ways that output all clauses of $\overline{\Gamma}$, but unfortunately in all known algorithms there may be exponential delay between the production of two consecutive clauses. There is no known *output-polynomial* algorithm for this problem.

There are many instances of such enumeration problems in the literature, for which no output-polynomial algorithm is known (despite the fact that, in contrast to NP-complete problems, it is trivial to output the first solution). The most famous one is to compute *all transversals of a hypergraph* (Eiter & Gottlob, 1995) (see the Appendix for a definition and discussion of this problem). As was pointed out by Eiter and Gottlob (1995), many enumeration problems arising in AI, databases, distributed computation, and other areas of Computer Science, turn out to be what we call in this paper *TRANSVERSAL-hard*, in the sense that, if they are solvable in output polynomial time, then the transversal problem is likewise solvable. It should be noted that recent research paints a rosier picture for the TRANSVERSAL problem, by showing that it can be done in *output-subexponential time* (Fredman & Khachiyan, 1996); but still, no output-polynomial algorithm is known.

**Theorem 1:** Enumerating all clauses of the Horn envelope of a given set $M$ of models is TRANSVERSAL-hard. As for the Horn core, selecting the Horn core (among the possibly exponentially many incomparable ones) with the maxim*um* number of models (i.e., the one that best approximates $M$) is NP-complete; furthermore even approximating the maximum within any constant ratio is NP-complete. □

Proofs of these results are given by Kavvadias, Papadimitriou and Sideri (1993); a version of the second result, in a different model and cost criterion, was shown independently by Cadoli (1993).

The computational problems related to updates and belief revisions are in fact much harder. Let $\Gamma$ be a set of Boolean formulas, and let $\phi$ be another formula; $\phi$ will usually be assumed to be of size bounded by a small constant $k$. We want to compute a new set of formulas $\Gamma + \phi$ —intuitively, the result of *updating* or *revising* our knowledge base $\Gamma$ by the new information $\phi$. There are many formalisms in the literature for updating and revising knowledge bases. First, if $\Gamma \& \phi$ is satisfiable, then all (with the single exception of Winslett, 1988) approaches define $\Gamma + \phi$ to be precisely $\Gamma \& \phi$ (we often blur the distinction between a set of formulas and their conjunction). So, suppose that $\Gamma \& \phi$ is unsatisfiable.

1. In the approach introduced by Fagin, Ullman, and Vardi (1983), and later elaborated on by Ginsberg (1986), we take $\Gamma + \phi$ to be not a single set of formulas, but the set of all maximal subsets of $\Gamma$ that are consistent with $\phi$, with $\phi$ added to each.

2. We shall consider here a more computationally meaningful variant called WIDTIO — for "when-in-doubt-throw-it-out"— in which $\Gamma + \phi$ is the *intersection* of the maximal sets mentioned in (1).

3. The above approaches are syntactic, in that they define the updated formulas explicitly. The remaining approaches are semantic, and they define $\Gamma + \phi$ implicitly by its set of models $\mu(\Gamma + \phi)$, given in terms of the set of models of $\Gamma$, $\mu(\Gamma)$, and that of $\phi$, $\mu(\phi)$ —notice that, if $\Gamma \& \phi$ is unsatisfiable, these two sets are disjoint. All five approaches take $\mu(\Gamma + \phi)$ to be *the projection* of $\mu(\Gamma)$ on $\mu(\phi)$, the subset of $\mu(\phi)$ that is closest to $\mu(\Gamma)$ —and they differ in their notions of a "projection" and "closeness." In Satoh's (1988) and Dalal's (1988) models, the projection is the subset of $\mu(\phi)$ that achieves minimal distance from *any* model in $\mu(\Gamma)$ (in Dalal's it is minimum Hamming distance, in Satoh's minimal set-theoretic difference). In Borgida's (1985) and Forbus's (1989) models, the projection is the subset of $\mu(\phi)$ that achieves minimal distance from *some* model in $\mu(\Gamma)$ (in Forbus it is minimum Hamming distance, in Borgida's minimal set-theoretic difference). Finally, Winslett's (1988) approach is a variant of Borgida's, in which the "projection" is preferred over the intersection even if $\Gamma \& \phi$ is satisfiable.

Eiter and Gottlob (1992) embark on a systematic study of the complexity issues involved in the various formalisms for updates and revisions. They show that telling whether $\Gamma + \phi \models \psi$ in any of these approaches (this is known as the *counterfactual problem*) is complete for levels in the polynomial hierarchy beyond NP —that is to say, hopelessly complex, even harder than NP-complete problems. When $\Gamma$ and $\phi$ are Horn, and $\phi$ is of bounded size, Eiter and Gottlob (1992) show their only positive result (for adverse complexity results, even in extremely simple cases, in approaches 1 and 2, see Theorem 2 parts (i) and (ii) below): The problem is polynomial in the approaches 3–7. This seems at first sight very promising, since we are interested in updating Horn approximations by bounded formulas. The problem is that *the updated formulas cease being Horn* (part (iii)).

We summarize the negative results original to this paper as follows (see the Appendix for the proofs; point (iii) is an easy observation which we include for completeness):

**Theorem 2:** Computing $\Gamma + \phi$, where $\Gamma$ is a set of Horn formulas and $\phi$ is Horn formula:

(i) Is TRANSVERSAL-hard in the Fagin-Ullman-Vardi-Ginsberg (1983; 1986) approach.

(ii) Is $\text{FP}^{\text{NP}[\log n]}$-complete in the WIDTIO approach (that is, as hard as any problem that requires for its solution the interactive use of an NP oracle $\log n$ times).

(iii) May result in formulas that are not Horn in the model-based approaches. □

Regarding Part (ii), a coNP lower bound and an $\text{P}^{\text{NP}[\log n]}$ upper bound were shown by Eiter and Gottlob (1992). Liberatore (1995) shows that, unless the polynomial hierarchy collapses, Horn updates result in formulas with inherently exponential length.

## 3. Incremental Recompilation

We now describe our scheme for representing propositional knowledge in a manner that supports rapid approximate reasoning and minimum-change updates/revisions. At time $i$ we represent our knowledge base with two Horn formulas $\underline{\Gamma}_i$ and $\overline{\Gamma}_i$. We start the process by computing the Horn envelope and core of the initial formula $\Gamma_0$, incurring a start-up computational cost —alternatively, we may insist that we always start with a Horn formula. Notice that we are slightly abusing notation, in that $\underline{\Gamma}_i$ and $\overline{\Gamma}_i$ may not necessarily be the Horn envelope and core of some formula $\Gamma_i$; they are simply convenient upper (weak) and lower (strict) bounds of the knowledge base being represented.

When the formula is updated by the formula $\phi_i$, the new upper and lower bounds are as follows:
$$\overline{\Gamma}_{i+1} := \overline{\overline{\Gamma}_i + \phi_i},$$
$$\underline{\Gamma}_{i+1} := \underline{\underline{\Gamma}_i + \phi_i}.$$

Here '+' denotes any one of the update formalisms discussed (the effect of the update formalism on our scheme is discussed in Section 4). That is, the new upper bound is the Horn envelope of the updated upper bound, and the new lower bound is the Horn core of the updated lower bound. Obviously, implementing this knowledge representation proposal relies on computing the Horn envelopes and cores of updated Horn formulas. We therefore now turn to this computational problem.

To understand the basic idea, suppose that we want to update a Horn formula $\Gamma$ by a *clause* $\phi = (\neg x \vee \neg y)$. Let us consider the interesting case in which $\Gamma\&\phi$ is unsatisfiable, and therefore $\Gamma$ can be written as $\Gamma = x\&y\&\Gamma'$ for some Horn formula $\Gamma'$ not involving $x$ and $y$. Consider now any model $m$ of $\Gamma$; it is of the form $m = 11m'$, where $11$ is the truth values of $x$ and $y$, and $m'$ is the remaining part of the model. The models of $\phi$ that are closest to it (both in minimum Hamming distance and in minimal set difference, as dictated by all five approaches) are the two models $01m'$ and $10m'$. Taking the union over all models of $\Gamma$, as the formalisms by Borgida and Forbus suggest, we conclude that $\Gamma + \phi$, the updated formula, is $(x \neq y)\&\Gamma'$. It is easy to see that the Horn envelope of this formula is just $(\neg x \vee \neg y)\&\Gamma'$, while the Horn core is either $x\&\neg y\&\Gamma'$ or $y\&\neg x\&\Gamma'$ —we can choose either one of the two.

As we mentioned in the introduction, if the update is a Horn implication, such as $\phi = (x\&y \to z)$ with $\Gamma$ of the form $x\&y\&\neg z\&\Gamma'$, the upper and lower approximations are these: $\overline{\Gamma + \phi}$ is $(x\&y \leftrightarrow z)\&\Gamma'$, while $\underline{\Gamma + \phi}$ is $x\&(y \leftrightarrow z)\&\Gamma'$ or $y\&(x \leftrightarrow z)\&\Gamma'$. The generalization to arbitrary Horn formulas is obvious.

**Theorem 2:** The Horn envelope and core of the update of a Horn formula $\Gamma$ by $\phi$, such that $\phi$ is a single Horn clause and $\Gamma\&\phi$ is unsatisfiable, in any one of the five model-based update formalisms 3–7 above, can be computed in linear time.

**Proof:** First suppose $\phi = (\neg x_1 \vee \neg x_2 \vee \ldots \vee \neg x_k \vee x_{k+1})$. Then, we can express $\Gamma$ as $x_1 \& x_2 \& \ldots \& x_k \& \neg x_{k+1} \Gamma'$, where $\Gamma'$ depends only on $x_{k+2}, \ldots, x_n$. $\Gamma'$ can be obtained in linear time by simply substituting the values $x_1 = 1, x_2 = 1, \ldots, x_k = 1, x_{k+1} = 0$ into $\Gamma$. Any model $m$ of $\Gamma$ is of the form $11\ldots10m'$, where $m'$ is a model of $\Gamma'$. The closest models of $\phi$ to $m$ (both in Hamming distance and minimality of set difference) are these (where all omitted bits are 1s):

$$011\ldots110m', 101\ldots110m', 111\ldots100m', 111\ldots111m'.$$

However, these are the models of the formula $\Gamma + \phi = \Phi\&\Gamma'$ where

$$\Phi = (\neg x_1 x_2 \ldots x_k \neg x_{k+1} \vee x_1 \neg x_2 \ldots x_k \neg x_{k+1} \vee \ldots \vee x_1 x_2 \ldots \neg x_k \neg x_{k+1} \vee x_1 x_2 \ldots x_k x_{k+1}).$$

Hence, in all revision/update formalisms, $\Gamma + \phi = \Gamma' \& \Phi$. Since $\Gamma'$ is a Horn formula, we have that $\underline{\Gamma + \phi} = \underline{\Phi}\&\Gamma'$ and $\overline{\Gamma + \phi} = \overline{\Phi}\&\Gamma'$; we must therefore compute the envelope and core of $\Phi$. It is not difficult to see that the possible cores of $\Phi$ are the formulas $x_1 x_2 \ldots x_{i-1} x_{i+1} \ldots x_k (x_i \leftrightarrow x_{k+1})$ for $i = 1, \ldots, k$, and thus

$$\underline{\Gamma + \phi} = x_1 x_2 \ldots x_{i-1} x_{i+1} \ldots x_k \&(x_i \leftrightarrow x_{k+1})\&\Gamma'.$$

On the other hand any model of the envelope of $\Phi$ either has $x_1 = x_2 = \ldots = x_{k+1} = 1$ or it has $x_{k+1} = 0$ and at least one of $x_1, \ldots, x_k$ equal to 0, so we can write

$$\overline{\Gamma + \phi} = (x_{k+1} \leftrightarrow x_1 x_2 \ldots x_k)\phi\Gamma'.$$

If $\phi$ is a negative clause (i.e. there is no $x_{k+1}$) then similarly $\underline{\Gamma + \phi}$ can be

$$\neg x_1 \& x_2 \& x_3 \& \ldots \& x_k \& \Gamma' \text{ and } \overline{\Gamma + \phi} = \phi\Gamma',$$

or any such formula, with another one of $\{x_1, \ldots, x_k\}$ negated. $\square$

## 4. Discussion

Theorem 3 implies that incremental recompilation in the face of single Horn clause updates can be carried out very efficiently for in all model-theoretic formalisms, *except for Winslett's*. Can this scheme be efficiently extended to the case in which $\phi$ has several Horn clauses? We next argue that the answer is negative. In fact, suppose that $\phi$ is the conjunction of several *negative* clauses, with no positive literals in them, and that $\Gamma$ is of the form $x_1 \& \ldots \& x_k \& \Gamma'$, where $x_1 \ldots x_k$ are the variables appearing in $\phi$. Consider a model $11\ldots1m'$ of $\Gamma$; what is the closest in Hamming distance model of $\phi$? The answer is *the model that has zeros in those variables among $x_1 \ldots x_k$ which correspond to a minimum hitting set of the clauses* (considered as sets of variables). Recall that a minimum hitting set of a family of sets is a set that intersects all sets in the family and is as small as possible. Finding the minimum hitting set of a family is a well-known NP-hard problem (Papadimitriou 1993). Therefore,

GOGIC, PAPADIMITRIOU, & SIDERItelling whether the Horn envelope of the updated formula (in the Forbus and Dalal models, which use Hamming distance) implies $x_i$ is equivalent to asking whether $i$ is not involved in any minimum-size hitting set —an coNP-complete problem! We have proved:

**Theorem 3:** Computing the Horn envelope of the update of a Horn formula by the conjunction of negative clauses in the Forbus or Dalal formalisms is NP-hard. □

Notice, however, that this hardness result requires that the update $\phi$ involve an unbounded number of variables. *We conjecture that the Horn envelope and core of a Horn formula updated by any formula involving a fixed number of variables can be computed in polynomial time* in all five model-based update formalisms —although the polynomial may of course depend on the number of variables. In our view, this is an important and challenging technical problem suggested by this work. We know the conjecture is true in several special cases —for example, the one whose unbounded variant was shown NP-complete in Theorem 4— and we have some partial results and ideas that might work for the general case.

### 4.1 The Choice of an Update Formalism

Of the five model-based update formalisms, which one should we adopt as the update vehicle in our representation scheme? Besides computational efficiency (with respect to which there are very minor variations), there is another important desideratum: The property that $\underline{\Gamma}_i \models \overline{\Gamma}_i$ (that is, that the "upper and lower bound" indeed imply one another in the desirable direction) must be retained inductively.

**Definition:** Let '+' be a change formalism. We say that '+' is *additive* if for any formulas $A$, $B$ and $\phi$ the following holds: $(A \vee B) + \phi = (A + \phi) \vee (B + \phi)$

**Theorem 5:** If $\underline{\Gamma}_i \models \overline{\Gamma}_i$, and the update formalism used is additive, then $\underline{\Gamma}_{i+1} \models \overline{\Gamma}_{i+1}$.

**Proof:** Let $\underline{\Delta}_i$ be such that $\overline{\Gamma}_i = \underline{\Gamma}_i \vee \Delta_i$. Then, we have:
$\overline{\Gamma}_{i+1} = \overline{\Gamma}_i + \phi_i = \overline{(\underline{\Gamma}_i \vee \Delta_i) + \phi_i} = \overline{(\underline{\Gamma}_i + \phi_i) \vee (\Delta_i + \phi_i)}$
On the other hand:
$\underline{\Gamma}_{i+1} = \underline{\Gamma}_i + \phi_i \models \underline{\Gamma}_i + \phi_i \models (\underline{\Gamma}_i + \phi_i) \vee (\Delta_i + \phi_i) \models \overline{(\underline{\Gamma}_i + \phi_i) \vee (\Delta_i + \phi_i)}$
and therefore $\underline{\Gamma}_{i+1} \models \overline{\Gamma}_{i+1}$. □

Winslett's formalism satisfies the additivity condition. This is because, by definition, the set of models of $\Gamma + \phi$ under this formalism is the union over all models $m$ of $\Gamma$ of some function of $m$ (namely, the set of models of $\phi$ that are closest to $m$); hence, disjunction (that is, union of models) distributes over +. Unfortunately, Winslett's formalism is the only model-based formalism whose efficient implementation in the case of single Horn clause updates is left open by Theorem 3. As the following example demonstrates, the remaining four model-based formalisms, by treating exceptionally the case in which $\Gamma$ and $\phi$ are consistent, are not additive, and may lead to situations in which the lower bound may fail to imply the upper bound:

**Example:** Suppose that we start with this (non-Horn) formula: $\Gamma_0 = (x \vee y) \& (x \vee z) \& (y \vee \neg z) \& (\neg y \vee z)$. Then the Horn core and envelope may be $\underline{\Gamma}_0 = y \& z$ and $\overline{\Gamma}_0 = (y \vee \neg z) \& (\neg y \vee z)$. If we next update by the $\phi = \neg x \& \neg y$ and apply any of the four update/revision

30

formalisms other than Winslett's, we get $\underline{\Gamma}_1 = \neg x \& \neg y \& z$ and $\overline{\Gamma}_1 = \neg x \& \neg y \& \neg z$. This establishes that using our technique in any one of these four formalisms may result in an "upper bound" that fails to be implied by the corresponding "lower bound." □

The possibility of an upper bound that does not imply the lower bound is, of course, a major weakness of our scheme. Overcoming it is a very interesting open problem. The most satisfying (and, in our view, likely) way of overcoming it is by developing a polynomial-time algorithm for updating Horn formulas by clauses in Winslett's formalism.

### 4.2 Characteristic Model Approximation

Kautz, Kearns, and Selman (1993, 1995) introduced an interesting alternative way of representing Horn formulae, namely, *characteristic models*. Let $\Gamma$ be a Horn formula, and let $H$ be its set of models. It is easy to see that $H = H^*$, where $H^*$ is the smallest set that contains $H$ and is closed under component-wise multiplication (AND) of its models; that is, iff $h_1, h_2 \in H^*$ implies $h_1$ AND $h_2 \in H^*$. This raises the possibility of the following alternative representation of $H$: We represent it by a minimal set of models $C$ such that $C^* = H(= H^*)$. This was first proposed by Kautz, Kearns, and Selman (1993); they called this set $C$ the set of *characteristic models* of $H$, and they showed that it is exactly the set of all elements of $H$ that cannot be represented as the AND of any subset of $H$. There are Horn sets that can be represented much more succinctly by characteristic models than by formulae, but there are also examples showing the opposite. One definite advantage of the characteristic models representation is that it allows for polynomial-time abduction (Kautz, Kearns, and Selman 1993, 1995).

Our next result points out a disadvantage of the characteristic models over Horn formulae. This result also frustrates immediately the possibility that updates and revisions can be done efficiently through the characteristic models of the Horn core and envelope (for the proof see the Appendix):

**Theorem 6:** Unless P=NP, the set of characteristic models of the intersection of two Horn sets of models, $H_1 \cap H_2$, cannot be computed in polynomial time, given the characteristic models of $H_1$ and those of $H_2$. □

Finally, it turns out that there is a similar approach to Horn approximation based on 2SAT, that is, formulas with at most two literals per clause. Although such approximations are plagued with similar complexity impediments as their Horn counterparts, they also enjoy similar updatability properties as those we showed for Horn clauses in Theorem 3 — except that the complexity depends quadratically on the number of literals in the update; see Gogic (1996).

### 4.3 Open Problems

We presented in this paper a proposal for the problems of *approximate reasoning* and *revisions/updates*. Although we have seen that each of the constituent problems is largely intractable, our work provides a computationally feasible and otherwise plausible way of approaching the *combined* problem. Although our positive complexity results are confined to single-clause updates, as far as we know, this is the first computationally feasible such proposal.

The main technical open problem raised by our work is to find polynomial-time algorithms for computing the Horn envelope and core of any Horn formula when updated/revised (in any of the five formalisms, most interestingly in Winslett's) *by any bounded formula*. We conjecture that such algorithms exist.

Our approach responds to updates and revisions by producing approximations of the knowledge base which become, with new updates, more and more loose. Naturally, its practical applicability rests with the quality of these approximations, and their usefulness in reasoning. This important aspect of our proposal should be evaluated experimentally. A complementary way of evaluating the effectiveness of our scheme would be to apply it to well-studied situations and examples in AI in which reasoning in a dynamically updated world is well-known to be challenging, such as reasoning about action.

## Acknowledgements

We are indebted to Bart Selman for many helpful comments on a preliminary version of the manuscript, and to the reviewers for a careful and fair evaluation. A preliminary version of this work was presented in the 1994 AAAI Conference.

## Appendix

A *hypergraph* $H = (V, E)$ is a finite set of nodes $V$, together with a set of hyperedges $E$, where each $e \in E$ is a subset of $V$ with at least two elements. Thus, a graph is a hypergraph in which all hyperedges are of cardinality two. A *transversal* $t$ of a hypergraph is a minimal hitting set of the hyperedges of $G$, that is, a set of nodes that has a nonempty intersection with all hyperedges in $E$, and such that each proper subset is disjoint from some hyperedge. TRANSVERSAL is the following computational problem: Given a hypergraph, produce the set of all of its transversals. It is not known whether this problem can be solved in *output-polynomial time*, that is, in time polynomial in both the number of hyperedges and transversals. An enumeration problem is called TRANSVERSAL-hard if TRANSVERSAL can be reduced in polynomial time to it.

Finally, the complexity class $\text{FP}^{\text{NP}[\log n]}$ is the class of all functions that can be computed in polynomial time when given access to at most $O(\log n)$ times to an oracle that correctly answers 3SAT questions (or questions related to any other NP-complete problem).

**Theorem 2:** Computing $\Gamma + \phi$, where $\Gamma$ is a set of Horn formulas and $\phi$ is a Horn formula: (i) Is TRANSVERSAL-hard in the Fagin-Ullman-Vardi-Ginsberg (1983; 1986) approach. (ii) Is $\text{FP}^{\text{NP}[\log n]}$-complete in the WIDTIO approach (that is, as hard as any problem that requires for its solution the interactive use of an NP oracle $\log n$ times). (iii) May result in formulas that are not Horn in any one of the model-based approaches.

*Proof of Part* (i): Let $H = (V, E)$ be a hypergraph, where $V = \{1, 2, \ldots, n\}$ and $E = \{e_1, \ldots, e_m\}$. We first construct $\Gamma$ and $\phi$ in the following way: The set of variables will be $X = \{x_1, \ldots, x_n\}$. $\Gamma = \{g_1, \ldots, g_n\}$ consists of the formulas $g_i = x_i$ for $1 \leq i \leq n$. Finally, $\phi$ consists of all clauses $(\neg x_{i_1} \vee \ldots \vee \neg x_{i_{k_j}})$, where $e_j = \{i_1, \ldots, i_{k_j}\}$ is an edge in $E$.

**Claim 1:** If $t$ is a transversal of $H$ then $M = \{g_i : 1 \leq i \leq n, i \notin t\}$ is a maximal subset of $\Gamma$ consistent with $\phi$.

*Proof:* We need to prove that $M$ is consistent with $\phi$ and that adding any other $g_i$ will change that. Let $v = (v_1, \ldots, v_n)$ be a truth assignment to the variables in $X$ such that $v_i = 0$ if $i \in t$ and $v_i = 1$ otherwise. From the definition we see that $v$ satisfies all formulas in $M$. On the other hand, take any clause in $\phi$, say $C_e = (\neg x_{i_1} \vee \neg x_{i_2} \vee \ldots \vee \neg x_{i_k})$. Then, edge $e = (i_1, \ldots, i_k)$ in $E$ intersects $t$, say in vertex $i_1$, which means that $v_{i_1} = 0$ and therefore clause $C_e$ is satisfied by $v$. So, $v$ satisfies both $\phi$ and $M$ and therefore the two are consistent.

Suppose now we add a function $g_i$ to $M$. From the definition of $M$ we see that $i \in t$ and therefore there is an edge $e$ that does not intersect $t - \{i\}$ (because $t$ is a transversal). But this now implies that all variables in $C_e$ appear in $M \cup \{g_i\}$ which means that $\phi$ and $M \cup \{g_i\}$ are inconsistent. □

**Claim 2:** If $M = \{g_{i_1}, \ldots, g_{i_k}\}$ is a maximal subset of $\Gamma$ satisfying $\phi$ then $t = V - \{i_1, \ldots, i_k\}$ is a transversal of $H$.

*Proof:* We first prove that $t$ intersects all edges in $E$. Take an edge $e$ and look at the clause $C_e$ of $\phi$. We know that $C_e$ is consistent with $M$ which means that there is some $g_j = x_j$ that does not belong to $M$ while $\neg x_j$ is in $C_e$. Now, from the definition of $t$ and $C_e$ we see that $j$ belongs to both $t$ and $e$.

Let us now prove that $t$ cannot be reduced to any smaller set. Suppose that $t - \{i\}$ is a transversal of $H$ where $i \notin \{i_1, \ldots, i_k\}$. Then, from Claim 1 we see that $M' = \{g_i, g_{i_1}, \ldots, g_{i_k}\}$ is consistent with $\phi$ which contradicts the fact that $M$ is maximal.

¿From Claims 1 and 2 it follows that we have reduced the problem of finding the transversals of a hypergraph to the problem of finding all maximal subsets of $\Gamma$ consistent with $\phi$, which means that computing $\Gamma + \phi$ is TRANSVERSAL-hard in the Fagin-Ullman-Vardi-Ginsberg approach.

**Part** (ii): $\text{FP}^{\text{NP}[\log n]}$ is the class of problems solvable in polynomial time with a Turing machine that can ask $O(\log n)$ questions to an NP oracle. The class is equivalent to the class of problems solvable by a polynomial time machine that can ask a linear number of questions to an NP oracle, but all in parallel. A problem complete for this class is: Given $n$ Boolean formulas $F_1, \ldots, F_n$, compute a vector $v = (v_1, \ldots, v_n)$ such that $v_i = 1$ if and only if $F_i$ is satisfiable. It is easy to see that instead of Boolean formulas we can give $n$ instances of any NP-complete problem.

To compute an update in the WIDTIO approach, it is enough to take every formula $g$ of $\Gamma$ and to ask the following question:

Q: *Can we choose a subset of $\Gamma - \{g\}$, consistent with $\phi$ when viewed alone, but inconsistent with $\phi$ when enlarged by $g$?*

Since all formulas are Horn, these are $n$ questions in NP that can be asked independently (in parallel), and therefore our problem is in $\text{FP}^{\text{NP}[\log n]}$.

In order to prove $\text{FP}^{\text{NP}[\log n]}$-hardness, we will first prove that answering (Q) above is NP-hard.

**Claim 3:** Let $\Gamma = \{g, g_1, g_2, \ldots, g_n\}$ be a collection of Horn formulas and let $\phi$ be a Horn formula. Then, telling whether $g$ belongs to all maximal subsets of $\Gamma$ consistent with $\phi$ is coNP-complete.

**Proof:** The proof is by reduction from a problem we call PURE3SAT (sometimes called, in our view, misleadingly, MONOTONE SAT), defined next. Call a clause *pure* if it contains only positive or only negative literals. A Boolean formula in CNF is called *pure* if it contains only pure clauses. PURE3SAT is the problem of deciding whether a pure 3CNF formula is satisfiable; it is known to be NP-complete (e.g. Garey & Johnson, 1979).

Suppose $F$ is a pure 3CNF formula containing positive clauses $P_1, \ldots, P_r$ and negative clauses $N_1, \ldots, N_s$, over the set of variables $\{x_1, \ldots, x_n\}$. We introduce variables $X_0, X_1, \ldots, X_n, Y_1, \ldots, Y_r$ and define

$g = (\neg Y_1 \vee \neg Y_2 \ldots \vee \neg Y_r)$
$g_i = X_i \& (\wedge_{x_i \in P_j} Y_j)$ for $1 \leq i \leq n$
$\phi = \wedge_{(\neg x_i \vee \neg x_j \vee \neg x_k) \in F} (\neg X_i \vee \neg X_j \vee \neg X_k)$

We now prove that $F$ is satisfiable if and only if there is a counterexample for $g$.

Suppose that $v = (v_1, v_2, \ldots, v_n)$ is a satisfying assignment for $F$. We define $\Gamma' = \{g_i : v_i = 1\}$, and we claim that $g$ is inconsistent with $\Gamma'$. Since each positive clause $P_j$ of $F$ is satisfied, there must be at least one true $x_i$ in $P_j$, and therefore each $Y_j$ is a conjunct of $\Gamma'$. But $g$ states that at least one of the $Y_j$'s must be false. On the other hand, $\Gamma'$ alone is consistent with $\phi$, because $\phi$ essentially contains the negative clauses of $F$, and we know that the trth assignment satisfies these. So, $\Gamma'$ is consistent with $\phi$, yet it cannot be enlarged by $g$ and therefore $g$ does not belong to all maximal subsets.

Conversely, suppose that such a $\Gamma'$ exists. We define $v_i = 1$ if $g_i \in \Gamma'$, and by the same line of reasoning as in the previous paragraph we prove that $v = (v_1, \ldots, v_n)$ satisfies $F$, which concludes the proof of the claim.

Let us take now $n$ instances of PURE3SAT. For every instance $F_i$ we build $\Gamma^i$ and $\phi^i$ the way described in the lemma, over a new set of variables. Our $\Gamma$ will be the union of all $\Gamma^i$'s, and $\phi$ will be the conjunctio of all $\phi^i$'s. By updating $\Gamma$ with $\phi$ using the WIDTIO approach, we obtain a new formula that will contain $g^i$ if and only if $F_i$ is unsatisfiable. Therefore, we can compute the answers to all $n$ instances just by looking at the update, which shows that our problem is $\text{FP}^{\text{NP}[\log n]}$-hard.

**Part** (iii): Let $\Gamma = x \wedge y$ and $\phi = \neg x \vee \neg y$. Notice first that $\Gamma \wedge \phi$ is not satisfiable. The set of models for $\Gamma$ is $\{11\}$ while the set of models for $\phi$ is $\{00, 01, 10\}$. In all model-based approaches, the set of models of $\Gamma + \phi$ is equal to $\{01, 10\}$, which obviously cannot be represented by a Horn formula.

This completes the proof of Theorem 2. □

**Theorem 6:** Unless P=NP, the set of characteristic models of $H_1 \cap H_2$ cannot be computed in polynomial time, given the characteristic models of $H_1$ and those of $H_2$.

**Proof:** We shall establish this by proving that the following problem is NP-complete:

MAXMODEL: *Given two sets of models $M_1$ and $M_2$ that are the characteristic models representing Horn sets $H_1$ and $H_2$ respectively, and an integer $k$, determine whether there is a model in $H_3 = H_1 \cap H_2$ with more than $k$ 1s.*

If MAXMODEL is NP-complete, then obviously one cannot compute in polynomial time the set of characteristic models of $H_1 \cap H_2$, given the characteristic models of $H_1$ and of $H_2$, unless of course P=NP. The MAXMODEL problem is obviously in NP because we can simply guess a model $m$, check whether it contains at least $k$ ones, and then check whether it belongs to $H_1$ and $H_2$.

In order to prove NP-hardness of the problem, we show a reduction from NODE COVER. We are given a graph $G = (V, E)$ with $|V| = n$ and $|E| = s$, and an integer $\ell$, and we are asked whether there is a set of fewer than $\ell$ nodes that cover all edges. We define two sets of characteristic models $M_1$ and $M_2$, as follows: For every edge $e_r = (i, j) \in E$ we have in $M_1$ a model $h = (h_1, \ldots, h_{s+n})$ such that $h_r = h_{s+i} = 0$ while all other bits of $h$ are 1, as well as a model $h' = (h'_1, \ldots, h'_{s+n})$ such that $h'_r = h'_{s+j} = 0$ while all other bits of $h'$ are 1. In $M_2$ we also add for every $i$ between 1 and $s$ a vector $h = (h_1, \ldots, h_{s+n})$ such that $h_1 = \ldots = h_s = h_{s+i} = 0$, and all other bits are 1. Our result now follows from this claim:

**Claim 4:** Let $k = n - \ell$. $G$ has a node cover of size less than $\ell$ if and only if $H_3 = H_1 \cap H_2$ has a model with more than $k$ ones.

*Proof:* Notice that $H_2$ contains all vectors having 0 at the first $s$ positions and at least one 0 among the last $n$ positions. In order to obtain a vector in $H_1$ that also belongs to $H_2$ (and therefore to $H_3$) we need to take vectors in $M_1$ that will produce (under bitwise multiplication) a vector having 0 on the first s positions. It is easy to see the idea behind our construction: If vector $h$ in $H_1$ has $h_r = h_{s+i} = 0$, it represents the fact that by putting node $i$ in the cover set edge $r$ is covered. So, if the set of vectors in $M_1$ produces under bitwise multiplication a vector in $H_2$, we have obtained a node cover in $G$, where a 0 at position $s + i$ means that node $i$ is in the node cover. Both directions of the claim are now obvious.

To return to the proof of Theorem 6, it follows from that it is NP-hard to find a maximal model of the intersection of two sets in the characteristic models representation. On the other hand, every maximal model belongs to the set of characteristic models, and therefore if we could find the intersection of two sets in the characteristic models representation we would be able to find the maximal model, completing the proof of Theorem 6. □